\documentclass{article}

\usepackage[final]{corl_2020} 
\usepackage{graphics} 
\usepackage{epsfig} 
\usepackage{mathptmx} 
\usepackage{times} 
\usepackage{amsmath} 
\usepackage{amssymb}  
\usepackage{color}
\usepackage{wasysym}
\usepackage{url}
\usepackage[toc,page]{appendix}

\usepackage[font=small,labelfont=bf]{caption}

\title{Towards Robotic Assembly by Predicting Robust, Precise and Task-oriented Grasps }

\author{
  Jialiang (Alan) Zhao\\
  The Robotics Institute, Carnegie Mellon University\\
  Pittsburgh, PA, United States\\
  \texttt{alanjz@cmu.edu} \\
  \And
  Daniel Troniak \\
  National Robotics Engineering Center \\
  Pittsburgh, PA, United States\\
  \texttt{dtroniak@nrec.ri.cmu.edu} \\
  \AND
  Oliver Kroemer \\
  The Robotics Institute, Carnegie Mellon University\\ 
  Pittsburgh, PA, United States\\
  \texttt{okroemer@andrew.cmu.edu} \\
}

\begin{document}
\maketitle


\begin{abstract}
Robust task-oriented grasp planning is vital for autonomous robotic precision assembly tasks.
Knowledge of the objects' geometry and preconditions of the target task should be incorporated when determining the proper grasp to execute.
However, several factors contribute to the challenges of realizing these grasps such as noise when controlling the robot, unknown object properties, and difficulties modeling complex object-object interactions. 
We propose a method that decomposes this problem and optimizes for grasp robustness, precision, and task performance by learning three cascaded networks.
We evaluate our method in simulation on three common assembly tasks: inserting gears onto pegs, aligning brackets into corners, and inserting shapes into slots.
Our policies are trained using a curriculum based on large-scale self-supervised grasp simulations with procedurally generated objects.
Finally, we evaluate the performance of the first two tasks with a real robot where our method achieves $4.28$mm error for bracket insertion and $1.44$mm error for gear insertion. 

\end{abstract}

\section{Introduction}
\label{sec:1}
\begin{figure}[ht]
    \centering
    \includegraphics[width=0.7\textwidth]{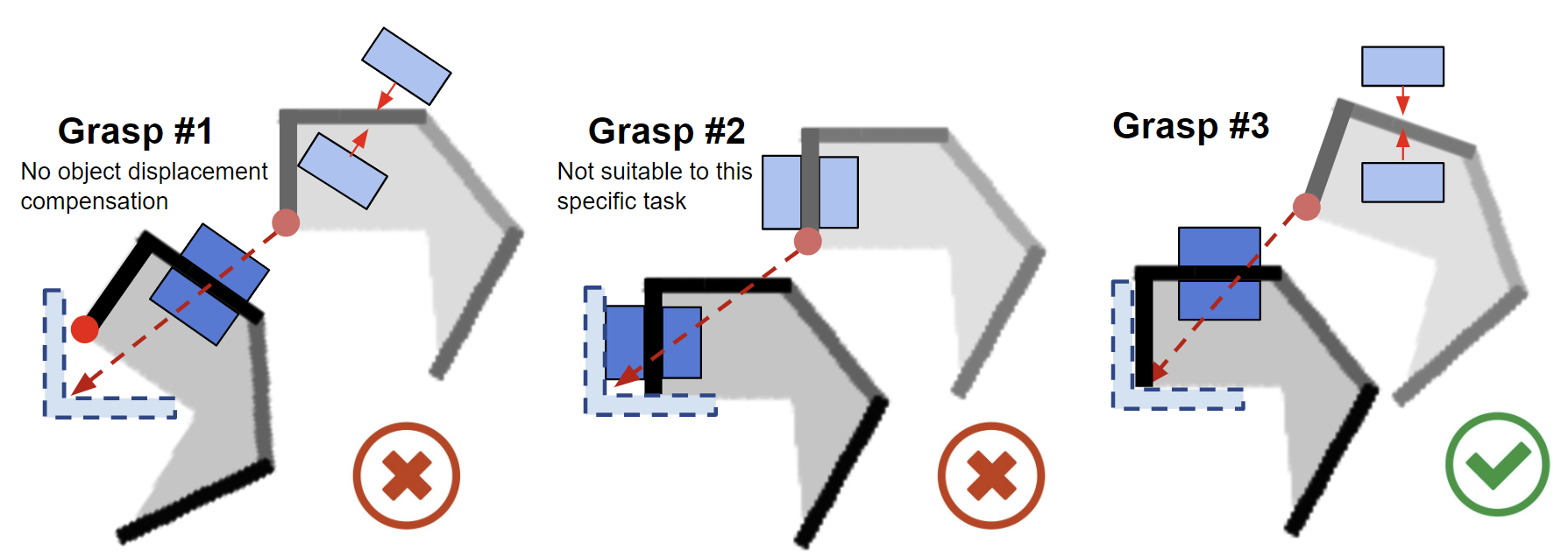}
    \centering
    \caption{\textbf{Grasps for a bracket alignment task} The red dot denotes the insertion frame that should be aligned with the corner. Attempt 1 does not consider the post-grasp object displacement and results in large insertion error (left), attempt 2 doesn't consider the insertion parameters and the gripper interferes between the object and the corner (middle), attempt 3 takes into consideration both displacement and task parameters. It achieves higher precision during insertion (right).}
    \label{fig:task_oriented_grasps}
\end{figure}

Grasping is a fundamental skill required by robots in autonomous manufacturing.
In robotic assembly, an agent needs to first grasp the object in some specific manner, then insert it into an assembly.
Precise and task-appropriate grasps are crucial to the success of assembly tasks.
For example, to insert a bracket into a corner, a good grasp should be aware of the possible post-grasp object displacement, and avoid being too close to the insertion point (see Fig.\ref{fig:task_oriented_grasps}).
The robot needs to have an estimate of how the object will be displaced as a result of the grasp, and how suitable this grasp is to the down-stream tasks.
A suitable grasp should constrain the in-hand pose of the object, but at the same time it should not interfere in the assembly task.
Such a grasp increases the chance of successful task completion.


Achieving such grasps is challenging due to the uncertainties in sensing, actuation, and object properties during interactions.
Recent works have largely focused on efficiently picking up objects without optimizing object in-hand poses~\cite{pinto2016supersizing,levine2018learning,jang2017end,mahler2019learning,fang2018learning,bousmalis2018using,fang2018multi}, or grasping objects in a specific location that is appropriate for manipulation tasks without considering precision~\cite{fang2018learning,thomas2018learning,luo2019reinforcement}.
Such grasps are generally unaware of task-specific contact limitations or post-grasp object displacement caused by the grasping motion.

In this work, we address the challenge of predicting task-oriented grasps for three example tasks: inserting gears onto pegs, aligning brackets into corners, and inserting shapes into slots. 
We propose a method that decomposes it into three sub-problems: choosing robust grasps, estimating grasp precision, and choosing a suitable grasp for the task.
We train a neural network for each of the sub-problems.
They are then cascaded together to predict the best overall grasp.
This grasp can then be used to achieve better precision in down-stream tasks. 
We train the policy with simulated grasps and we show the efficiency of our proposed approach both in simulation and on a real robot.
\section{Related Works}
\label{sec:2}
Researchers have made many contributions on planning robust robotic grasps.
A large number of them have focused on using analytical approaches based on physical models, such as resisting external wrenches~\cite{ferrari1992planning}, constraining the object's movement~\cite{rodriguez2012caging,dogar2011framework}, or combining multiple analytical metrics~\cite{rubert2018characterisation,leon2014characterization}. 
However, they usually assume the object is static and its pose is known. 
Other works have used data-driven grasp synthesis to predict grasps directly from sensory inputs. While early works used human-labeled grasps~\cite{jiang2011efficient,lenz2015deep}, an increasing number of works have started to collect data in a self-supervised manner by attempting thousands of grasps physically in the real world~\cite{pinto2016supersizing,levine2018learning,jang2017end} or in simulation~\cite{mahler2019learning,fang2018learning,bousmalis2018using,fang2018multi}.
However, most of those grasp planners only focus on efficiently picking objects up.
They do not take into account down-stream tasks that require objects to be grasped in a specific manner or with a precise in-hand pose.

Due to imperfectly approximated dynamics, as well as noise in sensing and actuation, a grasp that a robot executes often results in object shifting during finger closure. 
Recent works have also studied how to avoid or decrease such error. 
Chen et al. explored a closed-loop active planning framework to improve precision for objects with known models ~\cite{chen2015uncertainty}.
They further generalized it to unseen objects by proposing a probabilistic signed distance function representation of object surfaces from depth images and analytical metrics~\cite{chen2018probabilistic}.
Dogar et al. used pushing to funnel the clutter of objects in front of grasps to reduce object pose uncertainty~\cite{dogar2011framework}.
Jentoft et al. analyzed the source of uncertainties in grasping tasks and predicted grasps that were robust to local variations based on the range of variations a grasp can tolerate~\cite{jentoft2018think}.
Our previous work reasoned about grasping precision to improve accuracy of unconstrained tasks by approximating post-grasp object displacement as a probabilistic distribution~\cite{zhao2019towards}.
Gupta et al. explicitly modeled the latent noise in perception and actuation~\cite{gupta2018robot}.
However, those works only constrain the errors introduced by task-agnostic grasps.
There is still no guarantee whether a grasp is suitable for the down-stream tasks.
 
Recent works have extended grasp planning for task-oriented grasping, where the planner also optimizes for the success of downstream tasks.
Dang et al. manually labeled grasps with semantics and learned a \textit{semantic affordance map} to relate object geometry to predefined grasps that are appropriate to different tasks~\cite{dang2012semantic}.
Gao and Tedrake transferred the semantic grasping problem into a key point detection and object reconstruction problem by learning an object representation based on semantic key points~\cite{gao2019kpam}.
Fang et al. proposed a method that was jointly optimized for grasping and manipulation skills with simulated grasps and generated objects~\cite{fang2018learning}.
There have also been works that use reinforcement learning (RL) methods with object geometry features extracted from CAD~\cite{thomas2018learning}, auxiliary sensing~\cite{luo2019reinforcement}, or classical controllers and demonstrations~\cite{schoettler2019deep}.
Those works take the constraints introduced by down-stream tasks into consideration.

In this work, we address the challenges of autonomous assembly where a grasp should be accurate and well-suited to the task. 
Our work stands in contrast with previous works by decomposing this problem into three sub problems: planning robust grasps, estimating post-grasp displacement, and choosing suitable grasps for tasks. 
We train one CNN for each, and cascade them to find the most appropriate grasp. We form it as a curriculum learning~\cite{bengio2009curriculum} problem to improve training performance. 
In this manner, the grasp planner can choose grasps that are robust, precise, and appropriate to the task.
The closest works to ours are~\cite{fang2018learning, detry2017task}, where they also learn task-oriented grasps and corresponding manipulation policies. 
However, our context is different, where ~\cite{fang2018learning} focuses mainly on utilizing tools to accomplish an interactive task which doesn't require high precision, and~\cite{detry2017task} focuses more on learning multiple grasps on an object that are suitable to different tasks.
\section{Learning Precise Grasping for Assembly}
\label{sec:3}

\subsection{Problem Statement}

Our goal is to enable robots to perform industrial insertion tasks with top-down, parallel-jaw grasps of singulated objects lying on a flat surface. 
Let the initial pose of an object on the work surface be $\mathbf{p}$.
We assume that during manipulation, an object can only undergo translational movements $(\Delta x,\Delta y,\Delta z)$ and planar rotation $\Delta \theta$. We define the post-grasp object displacement as $\Delta\mathbf{p}=[\Delta x, \Delta y, \Delta z, \Delta \theta]^T$.
We assume the robot will not be able to use additional sensors to perform in-hand localization.
The grasp planner needs to predict the displacement $\Delta\mathbf{p}$ based on object pose $\mathbf{p}$, candidate grasp $\mathbf{g}$, and observation $\mathbf{o}$. The observation $\mathbf{o}$ is a depth image of the object captured with a top-mounted stereo camera. 

A grasp $\mathbf{g}$ also has 4 degrees of freedom $\mathbf{g} = [g_x, g_y, g_z, g_\theta]^T$,
where the position $(g_x, g_y, g_z)\in \mathbb{R}^3$ is the location of the center of gripper relative to the object's geometric center $\mathbf{p}$.
The orientation parameter $g_\theta \in [-\frac{\pi}{2}, \frac{\pi}{2})$ denotes the planar rotation of the gripper.

We choose three commonly seen tasks in manufacturing sectors to evaluate our method on: inserting gears onto pegs, aligning brackets into corners, and shape insertion. 
In order to successfully accomplish those tasks, the robot needs to first perform a grasp that can robustly and reliable pick up the object. 
It also needs to have an accurate estimation of the post-grasp object pose, as the clearance between the object and the assembly can be small. 
Furthermore, a good grasp should be suitable to the down-stream manipulation, i.e. it should not interfere during the assembling step, as illustrated in Fig.\ref{fig:task_oriented_grasps}.
The task parameters are denoted as $\mathbf{k}$. 
It encodes the task type and the transformation from the insertion frame to the object geometric center $T_{o}^{i}$.
We denote the insertion error as $\mathbf{\epsilon}$.
For the bracket alignment task, $\mathbf{\epsilon_b}$ is defined as the Euclidean distance between the insertion frame and the assembly.
The insertion frame is a pre-defined coordinated frame originated at a corner on a bracket, and aligned with the direction along which it should be inserted into the assembly.
For gear and shape insertion, $\mathbf{\epsilon_g}$ and $\mathbf{\epsilon_s}$ are defined as the negative ID of the best peg or slot that could be successfully inserted onto. 
Higher ID means smaller clearance. 
We define 6 pegs with diameters $\diameter\{3, 4, 4.5, 5, 5.5, 6\}$mm as Peg \#1-6,
as well as 4 slots with clearance $\{9, 6, 4, 3\}$mm as Slot \#1-4.
Our goal is to develop a planner that predicts high quality grasps that lead to high accuracy insertion. 
Such a planner should also generalize over novel objects that are unseen during training.

\subsection{Overview of Approach}

The proposed method divides the grasp planning problem into three sub-problems: (1) predicting the task-agnostic grasp quality in terms of picking up the object, (2) predicting the post-grasp object displacement, and (3) predicting the insertion quality of a grasp based on the task parameters.
Individual convolutional neural networks were trained to predict grasp quality on each subtask.
They are then cascaded to gradually funnel down randomly generated candidate grasps to the best ones.

\textbf{Grasp Quality Prediction}:
The grasp quality $Q_G$ of a grasp $\mathbf{g}$ with observation $\mathbf{o}$ is defined as the probability of a successful lift $S$ using the grasp, $Q_G(\mathbf{g}, \mathbf{o}) = P(S = 1 | \mathbf{g}, \mathbf{o})$. 
We learn this mapping $Q_G(\mathbf{g}, \mathbf{o})$ using a CNN that we refer to as the Grasp Quality Network (GQN).

\textbf{Grasp Displacement Prediction}:
Let $\Delta\mathbf{p} = [\Delta\mathbf{x}, \Delta\mathbf{y}, \Delta\mathbf{z}, \Delta\mathbf{\theta}]^T$ denote the post-grasp displacement of an object. 
We learn a CNN to predict $\Delta\mathbf{p}(\mathbf{g}, \mathbf{o})$ from grasp $\mathbf{g}$ and observation $\mathbf{o}$, which we refer to as the Grasp Displacement Network (GDN). 

\textbf{Task Quality Prediction}:
We define the insertion quality $Q_I$ of a grasp $\mathbf{g}$ with observation $\mathbf{o}$ and GDN estimated displacement $\Delta\mathbf{p}$ as the probability of a successful insertion with insertion error $\mathbf{\epsilon}$ smaller than a threshold $\mathbf{\sigma}$, under task parameter $\mathbf{k}$, $Q_I(\mathbf{g}, \mathbf{o}, \Delta \mathbf{p}, \mathbf{\sigma}, \mathbf{k}) = P(\mathbf{\epsilon} < \mathbf{\sigma} | \mathbf{g}, \mathbf{o}, \Delta \mathbf{p}, \mathbf{k}, \mathbf{\sigma})$. This mapping is learned with a CNN that we refer to as the Insertion Quality Network (IQN). 
Curriculum learning~\cite{bengio2009curriculum} has been proven to be able to outperform learning from randomly presented training data by gradually increasing the complexity of presented data during training.
We form IQN as a curriculum learning problem by gradually decreasing the insertion error tolerance $\mathbf{\sigma}$.

Given the three learned networks, we form a grasp planner that first chooses grasps with high probability of successfully picking up the object, then estimates the post-grasp object displacement for each grasp, and finally filters them with the probability of completing the insertion task.

\vspace{-2mm}

\subsection{Self-Supervised Data Collection in Simulation}
To generate grasps for training the networks, we first procedurally generate $5,000$ object models for each of the tasks. 
We then collect a total of $520,000$ simulated data points for training the task-agnostic grasp quality network (GQN) and the displacement estimation network (GDN).
After GQN and GDN are trained, we collect another $200,000$ simulated data points for training the insertion quality network (IQN).

\subsubsection{Procedural Generation of Brackets and Gears}
\begin{figure}[!ht]
    \centering
    \includegraphics[width=\textwidth]{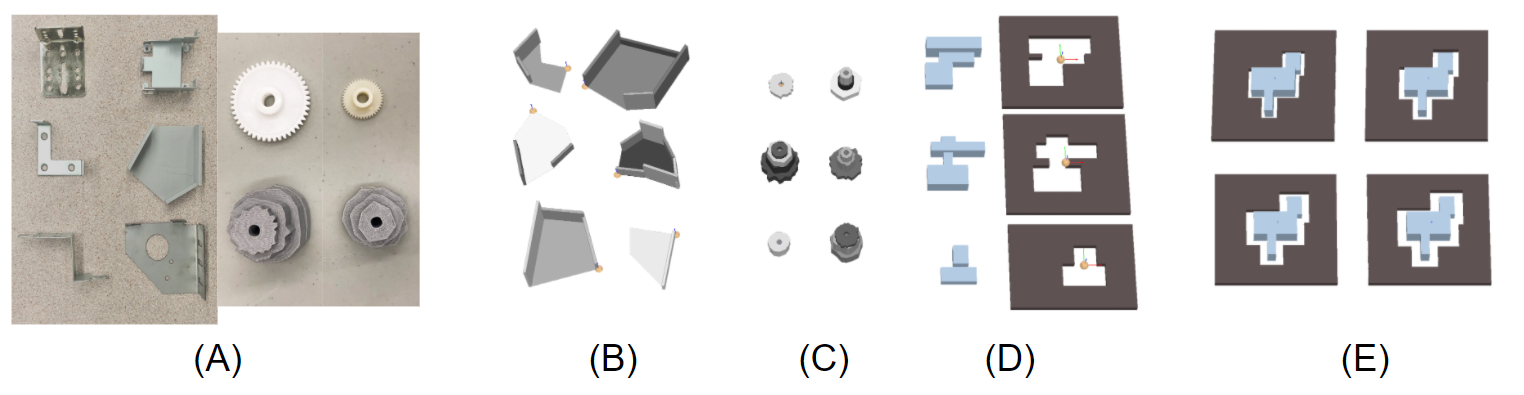}
    \centering
    \caption{\textbf{Parts and assemblies used in training and evaluation} (A) shows the actual brackets and gears that we use in real world manufacturing tasks; (B-D) show brackets with polygon base and walls, multi-layer gears with $\diameter6.3$mm shaft holes, as well as shapes and slots for shape insertion. We generate 4 slots for each of the shape, with different clearance, as shown in (E). The shape insertion task is only conducted in simulation.}
    \label{fig:sample_objects}
\end{figure}

In order to generalize our method to objects unseen during training, we adopt the idea of domain randomization by training our models with a large number of different objects.
However, the existing 3D model datasets do not contain enough parts suitable for the insertion tasks while exhibiting rich variations in terms of their geometric and physical properties. 
We generate a large set of diverse and realistic objects that are close to the actual parts with primitive shapes for each of the tasks.
Some examples are shown in Fig.\ref{fig:sample_objects}. 

\textbf{Bracket Generation}
The generated brackets are constructed with two components: a flat polygon base plate with 4 to 8 vertices and 1 to 5 walls with random heights $\in[10, 50]$mm. 
The base plate can be convex or concave, and we make sure that there is always at least one $90\deg$ corner, which we mark as the insertion frame.

\textbf{Gear Generation}
A generated gear contains up to 5 layers, and the radius of one layer is always smaller than  the layer under it. All the gears have a shaft hole of $\diameter6.3$mm such that all of them are able to be inserted onto the largest peg ($\diameter6$mm) that we test with. 
Each layer is approximated with a polygon of 4 to 50 vertices. More vertices make a layer smoother and rounder.

\textbf{Shape and Slot Generation}
A shape is composed with 1 to 5 cuboids. Each of the cuboid has length $\in [10, 70]$mm, width $\in [20, 30]$mm, and height $\in [10, 30]$mm. 
Adjacent cuboids are guaranteed to be connected to each other.
For each of the shape, we generate 4 slots, with clearance $\in \{4, 6, 8, 10\}$mm. All the slots have the same height of 10mm.

\subsubsection{Simulated Grasps Generation}
We gather grasps using the robotic simulation framework V-REP~\cite{rohmer2013v} and PyRep~\cite{james2019pyrep} with Bullet $2.78$~\footnote{\url{https://pybullet.org/}} as the physics engine. 
We assume each object has uniform density.

\textbf{Data Generation for GQN and GDN}
An object-centric depth image is captured at the start of each grasp attempt. 
The robot executes a grasp $\mathbf{g}$, where the grasp center $(g_x,g_y,g_z)$ is uniformly sampled from the bounding box of the object, and the grasp orientation $g_\theta$ is uniformly sampled from $[-\frac{\pi}{2}, \frac{\pi}{2})$.
In contrast to previous works that sample antipodal grasps~\cite{mahler2019learning, mahler2017dex}, we use uniformly sampled grasps as using antipodal grasps may introduce a bias in the dataset when estimating post-grasp displacements.
For each grasp attempt $\mathbf{g}$, we record the lift success $S$ of the grasp, the overhead depth image $\mathbf{o}$, and the post-grasp object displacement in world frame $\Delta \mathbf{p}$.

\begin{figure}[!ht]
    \centering
    \includegraphics[width=0.8\textwidth]{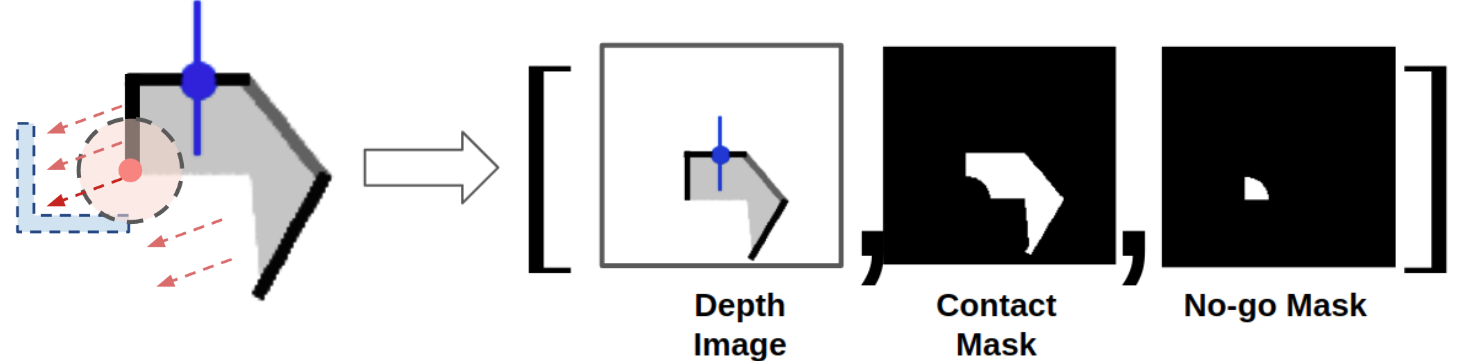}
    \centering
    \caption{\textbf{Contact and no-go masks generation} Both masks are binary, indicating the object areas $2cm$ away from or within $2cm$ to the insertion frame. We use binary masks to incorporate task parameters.}
    \label{fig:masks}
\end{figure}

\textbf{Data Generation for IQN}
After GQN and GDN are trained, we collected another $200,000$ data for training the insertion quality network, IQN. 
The robot first samples and executes a random grasp $\mathbf{g}$. 
If $\mathbf{g}$ successfully picks the object up, we use GDN to estimate the object displacement $\Delta \mathbf{p}$, which is used to correct the object in-hand pose.
The robot then uses this corrected pose to do the insertion task.
If $\mathbf{g}$ does not pick the object up, this trial will be aborted and it won't be used as training data to learn the IQN.
An insertion error $\mathbf{\epsilon}$ is recorded.
Besides capturing a depth image, we also calculate a binary contact area mask and a binary no-go area mask to encode the task parameters $\mathbf{k}$. The contact area mask is the object area that is $2$cm away from the insertion frame, while the no-go area is the object area within $2$cm to the insertion frame, as illustrated in Fig.\ref{fig:masks}.
An insertion frame is defined as a coordinate frame on the part which should be inserted into and aligned with the assembly.
For each of the task, it is located at (A) the shaft hole on gears, (B) the $90\deg$ corner on brackets, and (C) the center of mass on shapes.

\vspace{-1em}

\subsection{Network Training}

\begin{figure}[!htbp]
    \centering
    \includegraphics[width=\textwidth]{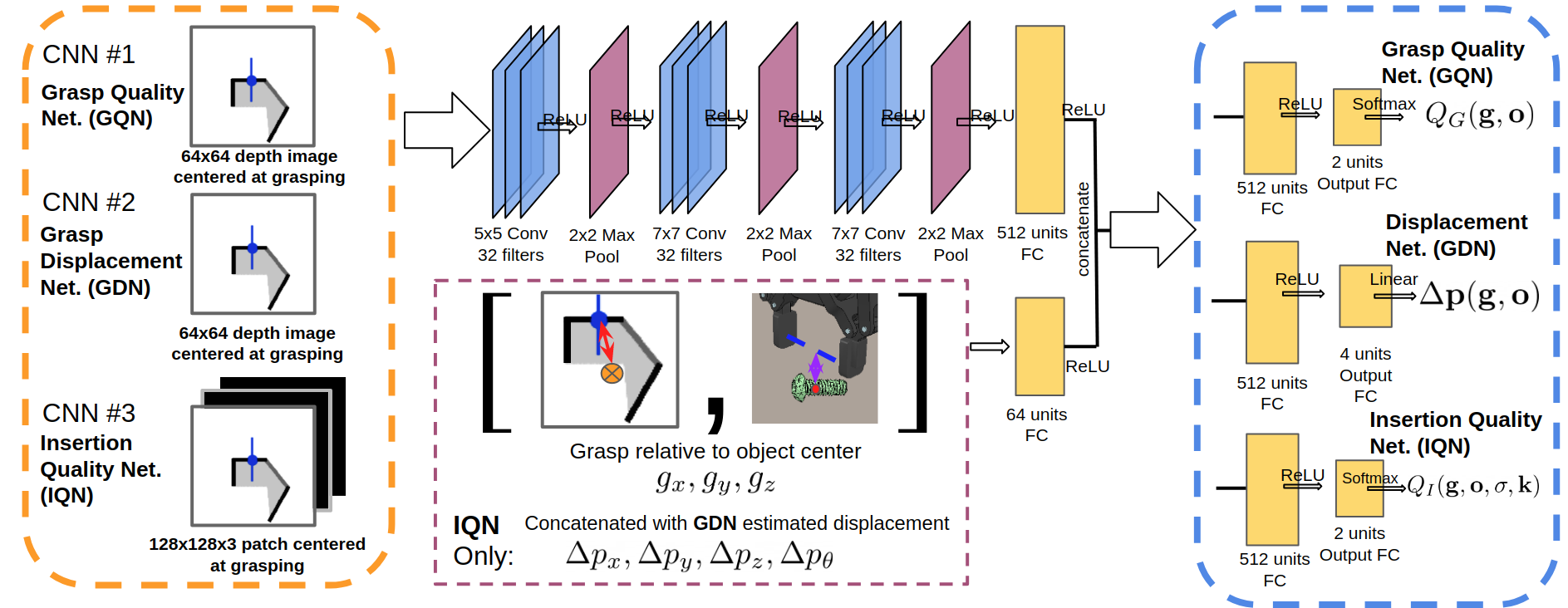}
    \caption{\textbf{Network structure} We trained three separate networks: the Grasp Quality Network (\textbf{GQN}), the Grasp Displacement Network (\textbf{GDN}), and the Insertion Quality Network (\textbf{IQN}). All the three networks have the same convolution layer architecture, but have different inputs and outputs. Note that although they share the same structure, they don't share weights.}
    \label{fig:networks}
\end{figure}

We train three types of CNNs - GQN, GDN, and IQN. While they all share a similar convolution architecture, they do not share weights. Rather, the GQN is trained first, and we use its learned convolution filters to initialize the filter weights of the GDN.
IQN has a different input dimension and is trained separately after GDN and GQN are trained. 
See Fig.\ref{fig:networks} for details on network architecture.

The Grasp Quality Network, $Q_G(\mathbf{g}, \mathbf{o})$, and the Grasp Displacement Network, $\Delta\mathbf{p}(\mathbf{g}, \mathbf{o})$ were trained using grasp-centric image patches, and the translation $(g_x,g_y,g_z)$ between the grasp and the object's geometric center. 
We train the GQN using a binary cross entropy loss and the GDN using a square error loss.
We train one GQN and one GDN and use them duing all the tasks.

The Insertion Quality Network $Q_I(\mathbf{g}, \mathbf{o}, \mathbf{\sigma}, \mathbf{k})$ was trained using grasp-centric image patches stacked with contact area masks and no-go area masks, as well as a 7 DOF vector composed of the grasp to object center translation $(g_x,g_y,g_z)$ and the GDN estimated displacement $(\Delta p_x, \Delta p_y, \Delta p_z, \Delta p_\theta)$.
IQN was trained using a binary cross entropy loss.
Because IQN is task-specific, we train one separate IQN for each of the three tasks.

We train IQN in curriculums.
Based on the distribution of collected data, we choose the insertion error thresholds for brackets to be $\mathbf{\sigma_b} = \{16.65, 8.32, 5.55, 4.16\}$mm and form it as a four-stage curriculum.
Thresholds for gears insertion are chosen as the negative IDs of the 6 pegs from the smallest one to the largest one, $\mathbf{\sigma_g} = \{-1, \cdots, -6\}$, which is formed as a six-stage curriculum.
Similarly, for shape insertion the thresholds are chosen as the negative IDs of the 4 slots $\mathbf{\sigma_s} = \{-1, \cdots, -4\}$. It is formed as a four-stage curriculum.
During training, IQNs are first fed with data from the stage which contains \textit{easier} data points (corresponds to a larger $\mathbf{\sigma}$). 
After convergence, they are then fed with data from the next stage, which contains \textit{harder} data points (a smaller $\mathbf{\sigma}$).
This process continues until the IQN is trained with data from all the stages.

\section{Experiments and Discussions}
\label{sec:5}
As our context is different from the other related works and there is no direct way to compare our method against them. We make an ablation study and compare our model with the following ablation baselines:

\begin{itemize}
    \item \textbf{GQN-only} - Select grasps only based on predicted grasp quality. Assume the grasped object has zero post-grasp displacement.
    \item \textbf{GQN+GDN} - Select grasps only based on predicted grasp quality. Adapt the insertion motion primitives based on post-grasp object displacement.
    \item \textbf{GQN+GDN+IQN} - Select grasps with both high predicted grasp quality and high insertion quality. Adapt the insertion motion primitives based on post-grasp object displacement.
\end{itemize}


\subsection{Evaluations in Simulation}
With the same object generation procedure, we generate another $500$ novel objects for each of the 3 tasks to perform evaluations on. 
We perform around $200$ simulated trials for each of the model. 


\begin{figure}[!htbp]
\centering
\includegraphics[width=\textwidth]{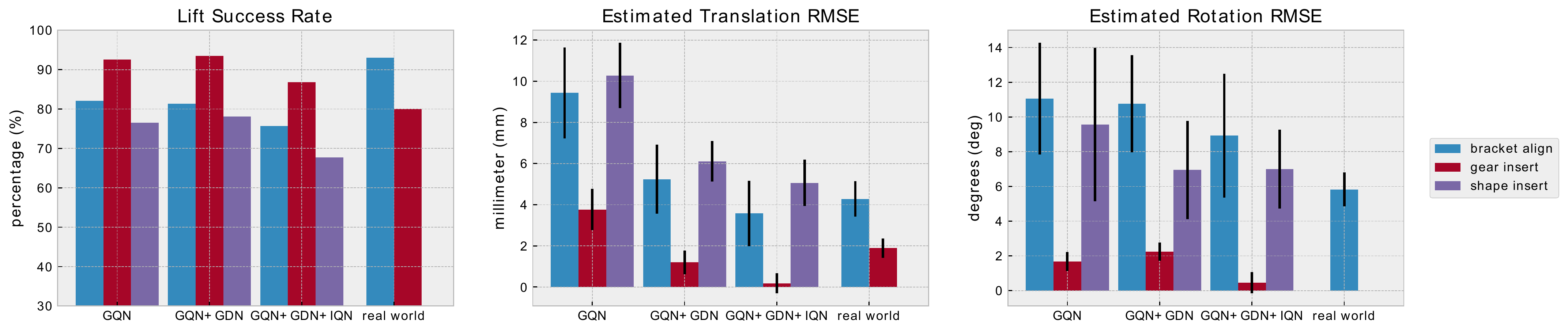}
\caption{\textbf{Lift success rate and RMSE of estimated object displacement} In real-world experiments, the lift success rate is $93\%$ for brackets and $80\%$ for gears; the translational RMSE's are $4.28$mm for brackets and $1.89$mm for gears; the rotational RMSE's are $5.83$deg for brackets. Real-world evaluation uses the \textbf{GQN+GDN+IQN} model.}
\label{fig:lift_success_displacement}
\end{figure}

\textbf{Lift success rate} is shown in Fig.\ref{fig:lift_success_displacement}. 
\textbf{GQN+GDN+IQN} has a slightly decreased lift success rate compared to the two ablation baselines, which means IQN has a negative effect in terms of lifting robustness.
We believe this is because the grasps that are suitable to tasks are sometimes less stable. 

\textbf{Displacement Estimation RMSE} is shown in Fig.\ref{fig:lift_success_displacement}. 
In our experiments, we observe that on some occasions it is non-trivial to tell how a grasp will displace an object even for humans, due to the partial geometry from a top-captured image and imperfect simulation dynamics.
GDN also struggles to produce reliable predictions in those occasions.
Although the GDN that we use in all the baselines is the same one, \textbf{GQN+GDN+IQN} has a consistently higher accuracy than \textbf{GQN+GDN}.
This is because the output from GDN also serves as an input to IQN.
IQN is used to rank grasps, and we believe it also implicitly learns whether a GDN estimation is reliable, or whether GDN can perform reliably under the current conditions.

\begin{figure}[!htbp]
\centering
\includegraphics[width=\textwidth]{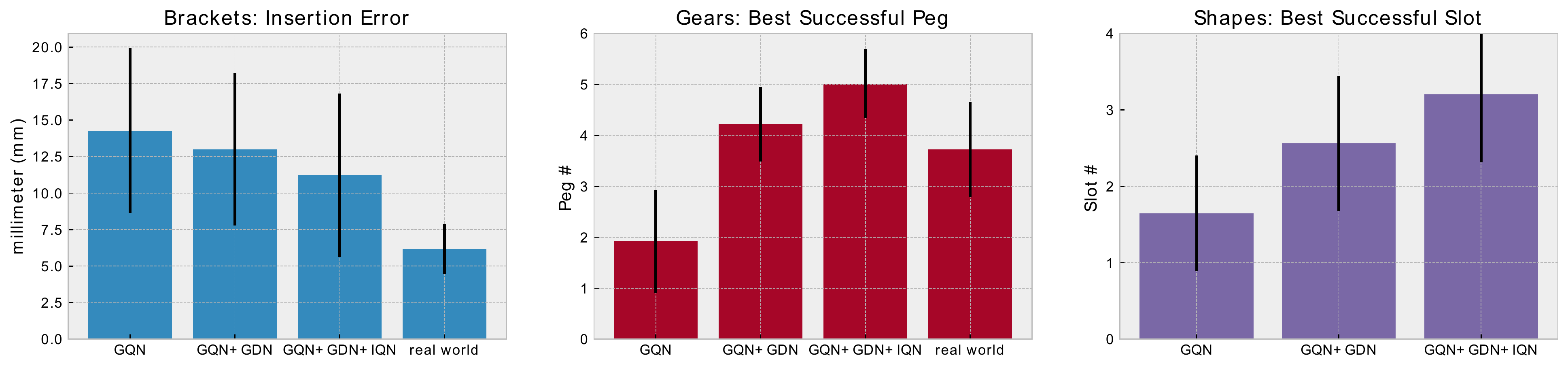}
\caption{\textbf{Insertion performance} Mean and std for bracket insertion error (left), best successful peg ID in gear insertion (middle), and best successful slot ID in shape insertion (right). For real-world experiments, the average bracket insertion error is $6.17$mm, the average best successful peg for gear insertion is $3.71$, or $1.44$mm insertion precision (calculated as shaft hole diameter subtracted by peg diameter).}
\label{fig:insertion}
\end{figure}

\begin{figure}[!htbp]
\centering
\includegraphics[width=\textwidth]{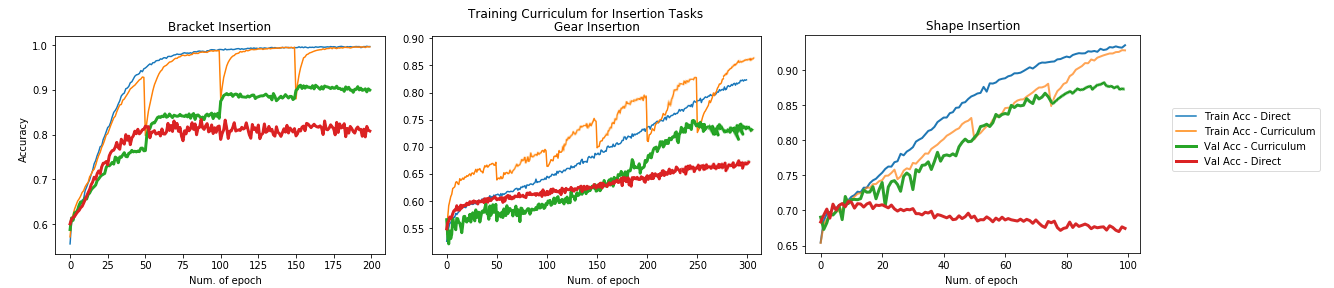}
\caption{\textbf{IQN training curves} Training as a curriculum outperforms training directly consistently.}
\label{fig:training_curve}
\end{figure}

\textbf{Insertion Performance} is shown in Fig.\ref{fig:insertion}, where \textbf{GQN+GDN+IQN} has the best performance. We notice that the simulator sometimes became unstable after making contact during bracket and shape insertion, which we believe is the cause for larger errors during those tasks. The validation accuracy when training the curriculums are shown in Fig.\ref{fig:training_curve}. 
In all three tasks, training IQN as a curriculum leads to higher accuracy and faster convergence than training directly.

\subsection{Evaluations with a Real Robot}
\begin{figure}[!htbp]
\centering
    \includegraphics[width=0.9\columnwidth]{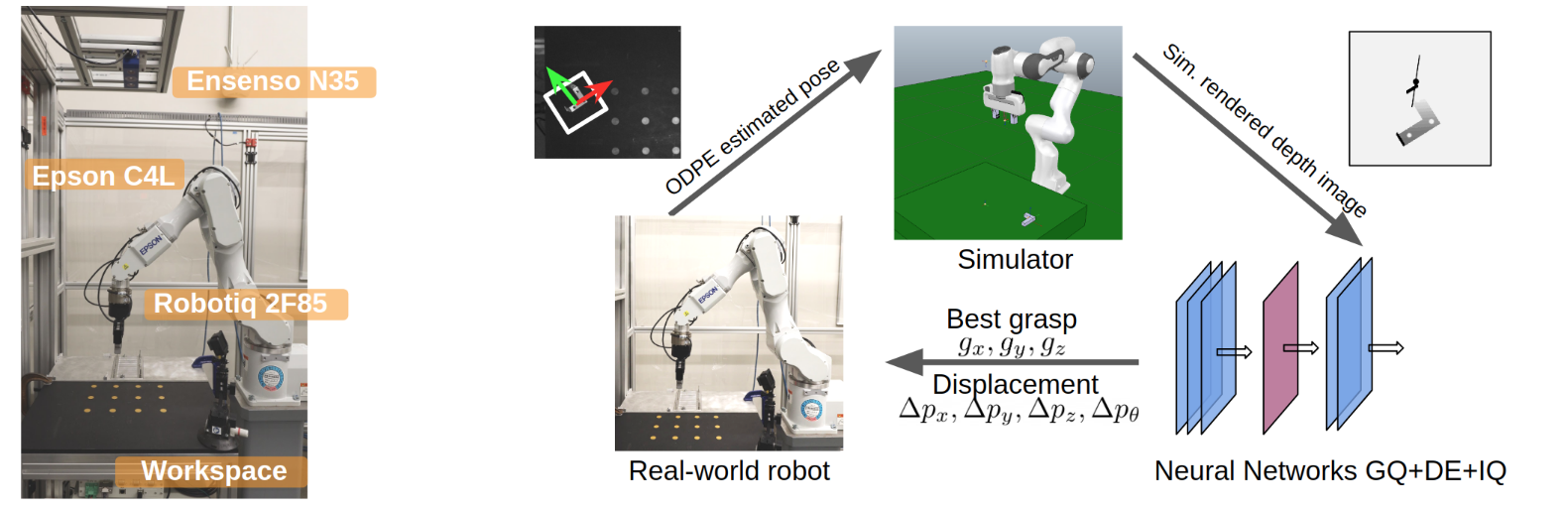}
    \caption{\textbf{Real-world experiment setup} (left) and \textbf{Planning pipeline in real-world experiments} (right). ODPE first estimates the pose of the object in the bin area, then it sends the pose to the simulator. The simulator sends the rendered depth image based on the estimated pose to the networks. The networks predict the best grasp $g_x, g_y, g_z$ and its corresponding displacement $\Delta p_x, \Delta p_y, \Delta p_z, \Delta p_\theta$, then sends the policy back to the robot.}
    \label{fig:real_world_setup}
 \end{figure}

The bracket alignment task and the gear insertion task are also evaluated with a real robot. 
We use a 7-DOF Epson C4L robot arm, with an over-head ENSENSO N35 stereo camera for depth sensing. 
We use 6 brackets (metallic and plastic) and 4 gears (resin and plastic) for evaluation. 
Five trials with each object are performed.
We use Epson Object Detection and Pose Estimation (ODPE) software to get the ground truth object pose. 
ODPE has a rated accuracy of 96\% that an object can be successfully detected and an object pose can be estimated within 5mm/5deg error. 
The setup is shown in Fig.~\ref{fig:real_world_setup}. The parts we used for experiments are shown in Fig.\ref{fig:sample_objects}.

To reduce the gap between simulation and reality we use the simulator to render depth images with the ODPE estimated object pose, instead of directly using real-world captured depth images. Appearance variations due to materials and lighting are thus removed. The pipeline is illustrated in Fig.\ref{fig:real_world_setup}. For more implementation details on the real-world evaluations, please refer to the appendices.

Our \textbf{GQN+GDN+IQN} model achieved a lift success rate of 93\% for brackets and 80\% for gears. 
The displacement estimation RMSE is $4.28$mm/$5.83$deg for brackets and $1.89$mm for gears. 
We are unable to get the rotational RMSE for gears because ODPE's orientation readings get unreliable when the parts are small and symmetrical.
The average task error for bracket insertion is $4.28$mm. 
The average best peg that the gears can be successfully inserted onto is $3.71$, or $1.44$mm insertion error.

\begin{figure}[!htbp]
    \centering
    \includegraphics[width=\columnwidth]{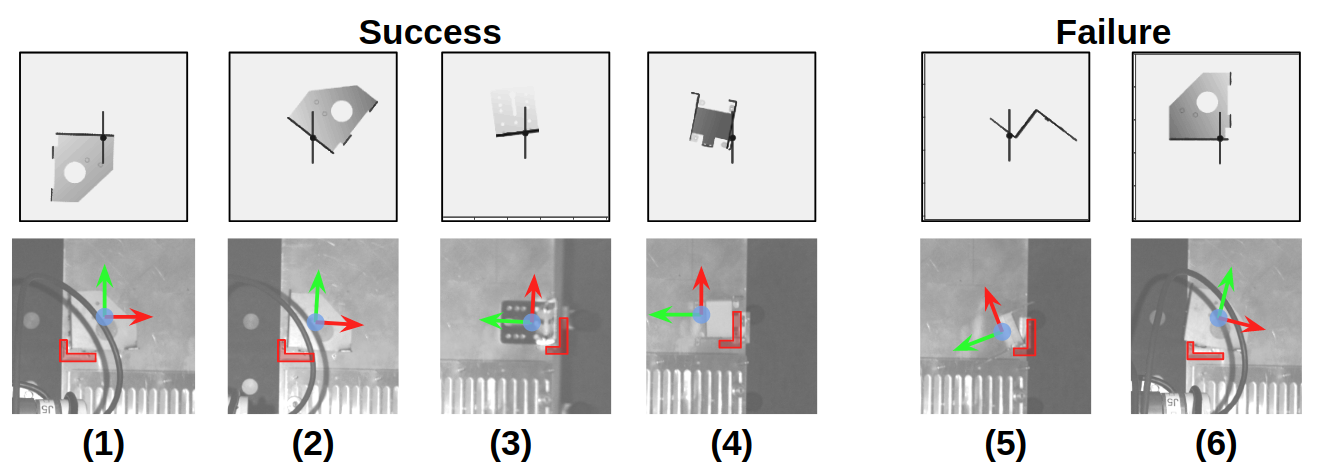}
    \caption{\textbf{Visualization of example grasps} Successful grasps should avoid the insertion frame and take into account of post-grasp object displacement (1-4), while (5) fails to estimate the correct post-grasp object displacement and (6) grasps too close to the insertion frame.}
    \label{fig:sample_grasps}
\end{figure}

In Fig.\ref{fig:sample_grasps} we visualize some grasps during real-world evaluations of bracket insertion.
There is an apparent gap between some of the real brackets and the procedurally generated ones, such as the holes in (1)(2) which do not exist in generated ones, or the more complex 3d structures in (4).
However, our model perform well even with those objects.
(1)(3) are examples where an initial grasp is already good enough, where (a) it is stable, (b) it does not incur huge object displacement and (c) it does not interfere in the insertion process. 
(2)(4) are example of grasps where (a)(c) still hold true, but incur substantial object displacement.
Those grasps are still good as long as the robot can correctly compensate the object's pose with the estimated displacement.
Here our planner produces accurate estimates and successfully accomplishes the task.
(5) is an example of grasps for which our model fails to produce reliable displacement estimation.
By reproducing it in simulation, we find that the displacement estimation is very close to the ground truth in simulation.
Thus we believe this is an issue mainly with the gap between the simulated and the real dynamics. 
(6) shows an example where a predicted grasp is not suitable for the insertion task.
In this example the grasp is too close to the insertion frame (right bottom corner of the bracket).
The gripper interferes between the bracket and the assembly, which makes accurate insertion not possible.

\section{Conclusion and Future Work}
We proposed a method to plan robust and precise task-oriented grasps by dividing the process into three sub-problems: grasp robustly, grasp precisely, and find grasps that are suitable for the task. 
We solve them by training one neural network for each, and chain them together to gradually funnel down candidate grasps. 
Estimation of task robustness of a grasp is formed as a curriculum learning problem.
We train the policy in simulation and transfer it to a real robot. 
Experimental results show that the proposed method efficiently improves the insertion performance, reducing the insertion error by 21\% for bracket insertion, 37.5\% for gear insertion and 35.7\% for shape insertion, compared with task-agnostic grasps without precision estimation.
We also showed that the learned policy successfully transferred to a real robot. 

In the future, our goal is to further improve and generalize our model by (1) introducing more realistic objects with diverse object properties such as mass density and friction coefficient, and (2) introducing other task constraints as well as experimenting with other tasks such as tool manipulation.  




\clearpage


\bibliography{corl}  

\clearpage
\begin{appendices}

\section{Real world evaluation details}

\subsection{Workspace configuration}

\begin{figure}[!htbp]
    \centering
    \includegraphics[width=0.75\textwidth]{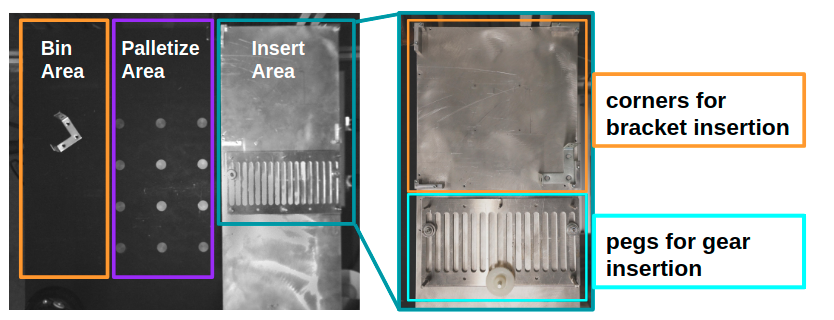}
    \caption{\textbf{Workspace configuration} The workspace is divided into a bin area, a palletize area and an insert area. The upper part of the insert area has 4 corners for bracket insertion; the bottom part has 6 pegs of diameters $\{\diameter3, \diameter4, \diameter4.5, \diameter5, \diameter5.5, \diameter6\}$mm for learning gear insertion curriculum.}
    \label{fig:workspace_config}
  \end{figure}

The workspace is divided into three areas: a bin area, a palletize area, and an insert area, as shown in Fig.\ref{fig:workspace_config}. The insert area is further divided into a bracket insert area and a gear insert area.
There are four metallic corners built in the bracket insert area. During experiments the robot picks one corner based on object's orientation.
There are six pegs built in the gear insert area. They have the same $\{\diameter3, \diameter4, \diameter4.5, \diameter5, \diameter5.5, \diameter6\}$mm diameters as in simulation.

\subsection{Evaluation workflow}

\begin{figure}[!htbp]
    \centering
    \includegraphics[width=\textwidth]{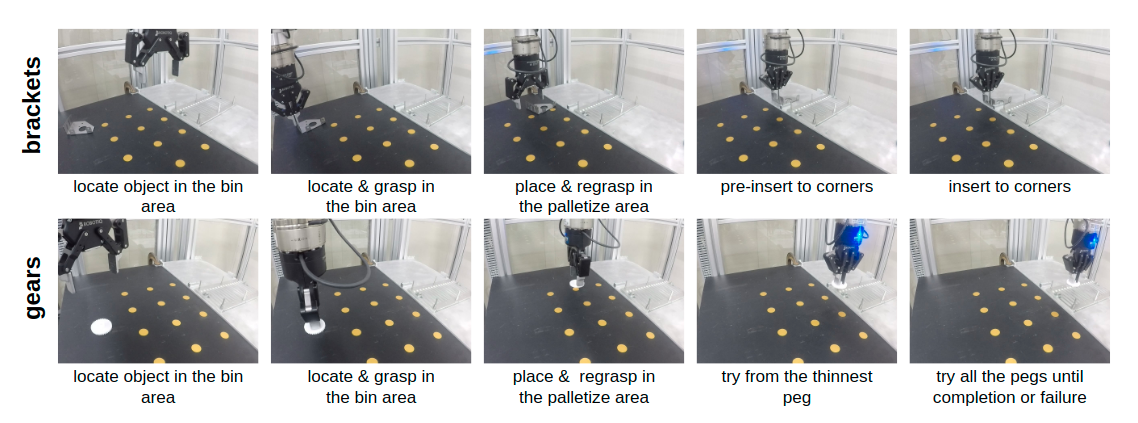}
    \caption{\textbf{Real-world experiment pipeline} Brackets insertion (up) and gears insertion (down).}
    \label{fig:exp_pipeline}
\end{figure}

The evaluation workflow (see Fig.\ref{fig:exp_pipeline}) is: (1) Human places one object in the bin area in a random pose. 
(2) The planner runs \textbf{GQN+GDN+IQN} networks to predict a best grasp and its corresponding displacement. The robot executes this grasp and compensates the displacement. 
(3) The robot places the object in the palletize area for pose estimation. The result is compared against the estimated displacement later during analysis. The robot then re-grasps the object using the same parameter as in placing.
(4) Brackets: the robot goes to a corner and prepare to insert. Gears: the robot goes to the smallest peg and tries to insert. 
(5) Bracket: insert it to the corner and record the insertion error. Gears: try all the pegs until completion or failure and record the best successful peg. 
After insertion, we run pose estimation again.
Finally the robot moves the object back to the bin area.

Note that the palletizing step is needed for estimating the in-hand object pose. We assume the pick and place in the palletize area do not change the in-hand object pose. 

\end{appendices}
\end{document}